%% file: main.tex
\pdfoutput=1

\documentclass[11pt]{article}

\usepackage[]{emnlp2021}

\usepackage{times}
\usepackage{latexsym}

\usepackage[T1]{fontenc}

\usepackage[utf8]{inputenc}

\usepackage{microtype}

\usepackage{graphicx}
\usepackage{subcaption}
\graphicspath{}
\usepackage{amsfonts}
\usepackage{amsmath}
\usepackage{cleveref}
\usepackage{algorithm}
\usepackage{algorithmic}

\input{macros}
\title{LADIS: Language Disentanglement for 3D Shape Editing}

\author{Ian Huang$^\dagger$,\hspace{1pt} 
    Panos Achlioptas$^\mathsection$,
    Tianyi Zhang$^\dagger$,  \\
    \bf{Sergey Tulyakov}$^\mathsection$, 
    \bf{Minhyuk Sung}$^\ddagger$ 
    {\normalfont and} \bf{Leonidas Guibas}$^\dagger$
    \vspace{0.5em}\\
    $^\dagger$Department of Computer Science, Stanford University \\
    {\tt \{ianhuang, tianyizhang, guibas\}@cs.stanford.edu}
    \vspace{0.5em}\\
    $^\ddagger$ School of Computing, KAIST
    \hspace{100pt}
    $^\mathsection $Snap Research\\
    {\tt mhsung@kaist.ac.kr}
    \hspace{10pt}
    {\tt \{panos, stulyakov\}@snap.com}
} 
  
\begin{document}
\maketitle

\input{sections/1_abstract}

\section{Introduction}

\input{sections/2_introduction}
\input{figures_latex/listening_architecture}

\section{Related Work}
\input{sections/3_related_works}

\section{Method}
\input{sections/4_method}

\section{Experiments}
\input{sections/5_experiments}

\section{Conclusion}
\input{sections/6_conclusion}

\section*{Limitations}
In this work, we study the task of language-based shape editing. 
We limit the scope of our study to previous datasets, where only English is studied.
We believe it's important future work to studying how other spoken and sign languages can be used for language-based shape editing.

\paragraph{Acknowledgements}
I. Huang and L. Guibas acknowledge the support of ARL grant (W911NF-21-2-0104), a Vannevar Bush Faculty Fellowship, and gifts from the Adobe and Snap Corporations.
M. Sung acknowledges the support of NRF grant (2022R1F1A1068681) funded by the Korea government(MSIT) and grants from Adobe, KT, Samsung Electronics, and ETRI.

\bibliography{anthology,custom}
\bibliographystyle{acl_natbib}
\pagebreak
\pagebreak
\appendix
\section{Appendix} \label{sec:appendix}
\input{sections/8_appendix}

\end{document}

%% file: sections/1_abstract.tex
\begin{abstract}
Natural language interaction is a promising direction for democratizing 3D shape design. However, existing methods for text-driven 3D shape editing face challenges in producing \emph{decoupled, local edits} to 3D shapes. We address this problem by learning disentangled latent representations that ground language in 3D geometry. To this end, we propose a complementary tool set including a novel network architecture, a disentanglement loss, and a new editing procedure. Additionally, to measure edit locality, we define a new metric that we call \emph{part-wise edit precision}. We show that our method outperforms existing SOTA methods by 20\% in terms of edit locality, and up to 6.6\% in terms of language reference resolution accuracy. Human evaluations  additionally show that compared to the existing SOTA, our method produces shape edits that are more local, more semantically accurate, and more visually obvious. Our work suggests that by solely disentangling language representations, downstream 3D shape editing can become more local to relevant parts, even if the model was never given explicit part-based supervision. 

\end{abstract}


%% file: sections/2_introduction.tex


Natural language has been used as a universally accessible and powerful interface to enable content creation and manipulation. For 2D images, Dall$\cdot$E~\cite{dalle2}, Imagen~\cite{imagen}, and StyleCLIP~\cite{styleclip} demonstrated impressive capabilities in creating and editing images using language descriptions.

However, understanding and grounding natural language in 3D geometry remains a challenge.
Recent efforts have all focused on using existing large pretrained vision-language models like CLIP~\citep{clip}, along with a differentiable renderer to interface 3D geometry with CLIP through 2D renderings. 
While these methods can perform geometry edits as described by language, they often also create unintended, additional, and unwanted changes. For example in \Cref{fig:teaser}, while the edit instruction asks for the chair legs to become thinner, existing methods can produce spurious editing artifacts elsewhere, like the back of the chair. To minimize the artifacts created by these changes, works such as ChangeIt3D \citep{changeit3d} often regularize their predicted edits to be very conservative, resulting in the overly subtle changes in regions of the shape where edits are requested, as shown in \Cref{fig:teaser}. Such edit behaviors can be counter-intuitive and detrimental to the editing experience, interfering with the design process.

\input{figures_latex/teaser}


%

We hypothesize that the source of non-local edits lies in the way we jointly represent edit descriptions and shapes.
To perform language-based shape edits, existing methods first learn a shape-language joint space using a classification or contrastive learning objective. 
Then, the system iteratively perform shape edits to maximize the similarity between the final shape representation and the edit instruction representation \cite{text2mesh, wang2022clip, hong2022avatarclip, jain2022zero}.
Problems arise when the joint representation space is entangled, i.e., the representations of independent edit instructions referring to unrelated shape edits have non-trivial co-dependencies. 
Optimizing with such entangled language representations can cause unintended correlations in unrelated shape edits.

In this work, we study how to achieve \emph{localized} language-based 3D shape editing through learning more disentangled language representations. Given an input shape and an utterance, the system should perform the correct edits to the shape while minimizing edits to parts of the shape that are unnecessary to shape integrity and undescribed by the edit description.



Thus, we aim to learn \textit{disentangled} grounded language representations that link language references to the correct parts of the shape. 
In an ideal disentangled latent space, \textit{independent} edit descriptions that describe mutually independent attributes and parts should be orthogonal to each other (e.g. ``the car has big wheels'' and ``the car has 2 doors''). \Cref{fig:listeneroverview} shows an example of two independent edit descriptions.


\input{figures_latex/listening_overview}

To this end, we propose a language disentanglement loss, named LADIS, and a multi-expert shape-difference encoder architecture. Whereas the LADIS loss enables the model to orthogonalize mutually independent utterances, the multi-expert shape-difference encoder allows the experts to specialize on different geometric features that span the semantics of the edit descriptions. These two design decisions enable our model to decouple different adjectives and part references within the edit descriptions, leading to improved disentanglement.

We note that a certain degree of geometry entanglement may be inherent in the integrity of the shape structure -- the use of different linguistic terms (e.g., ``table top'', ``table leg'') does not always imply perfect disentanglement of the part locations or geometries. For example, we typically expect co-dependencies between the extent of a four-leg table top and its leg locations. Such dependencies have to be learned from data and encoded in the shape space, making our task quite challenging.

\input{figures_latex/editing_overview}
Even with a more disentangled joint representation space, imperfect optimization makes shape editing difficult.
Empirically, we find that optimization-based 3D editing approaches may generate invalid objects. 
For example, failed shape edits can cause a chair shape to stop looking like a realistic chair.
To combat this issue, we additionally introduce two modifications to a standard optimization procedure: neighborhood simplex editing (NSE) and output-driven edit step-size adjustment (ODESSA). 
NSE re-frames shape editing as interpolating between a source shape and its nearest neighbors, which helps the edited shape to remain realistic.
ODESSA enables every optimization step to produce a constant amount of change in the output 3D space, allowing us to cope with metric ambiguities in language instructions. 
We evaluate our system on the ShapeTalk~\cite{changeit3d} and ShapeGlot~\cite{shapeglot} datasets.
To quantify the degree of edit locality, we propose \textit{part-wise edit precision} (PEP), a new metric that measures if shape edits are restricted to the parts that are mentioned by the edit description. 
Experimental results show that our system demonstrates an improved ability to resolve language references and successfully produce more local shape edits (20\% higher PEP than the closest competing system~\cite{changeit3d}). 
Additionally, human evaluation comprising of 500 annotations shows that compared to ChangeIt3D~\cite{changeit3d}, our method produces 3D shape edits with higher part-wise locality, semantic accuracy and visual obviousness.
Furthermore, our analysis of the language representations shows a higher level of disentanglement of shape part concepts and adjectives, demonstrating that disentangling language representation enables larger, more correct, and more localized shape edits, without ever equipping our model with explicit part supervision.
The code for LADIS can be found at \url{https://github.com/ianhuang0630/LADIS}.

%% file: figures_latex/teaser.tex
\begin{figure}
  \centering
  \includegraphics[width=\columnwidth]{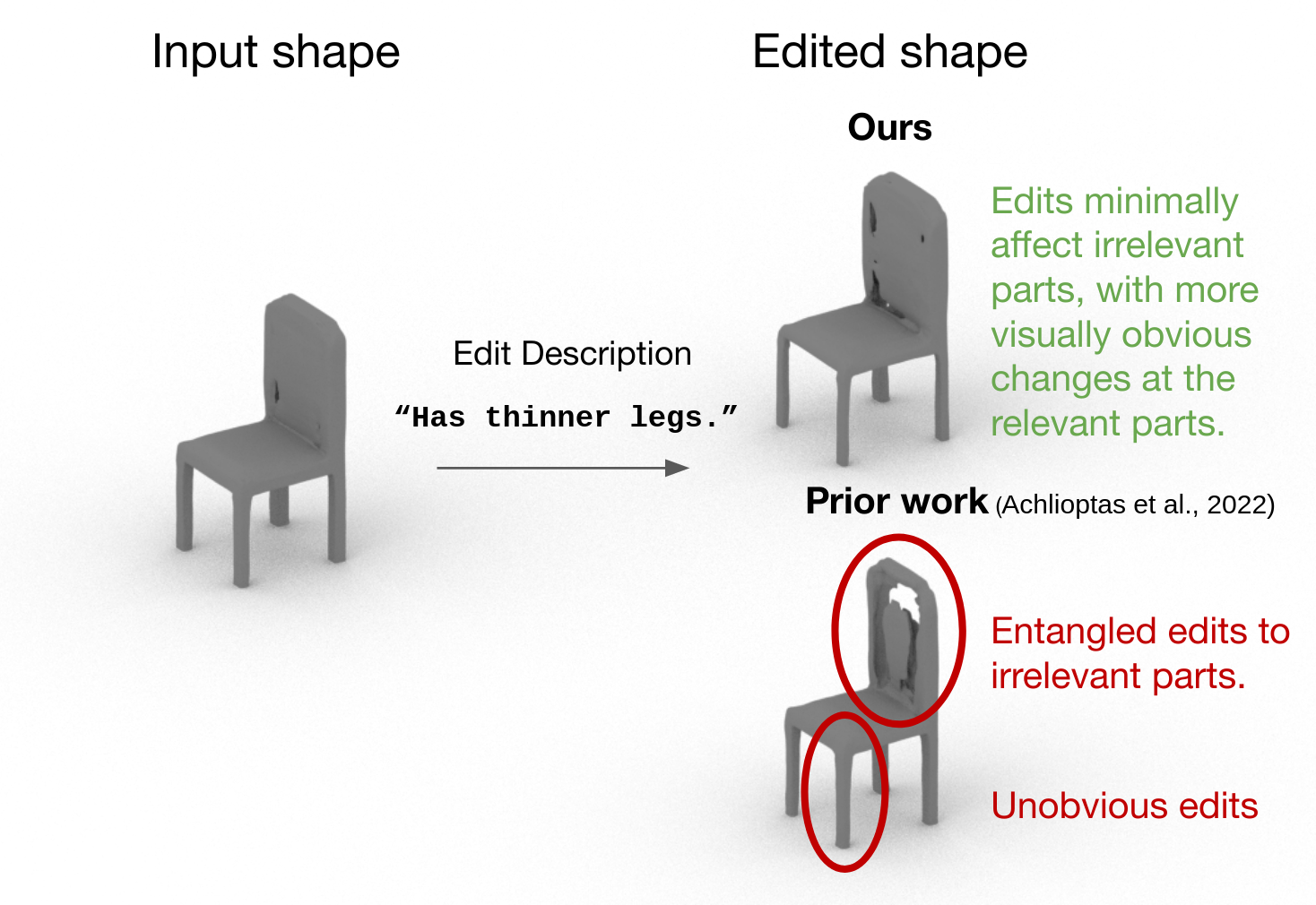}
  \caption{An example of the language-conditioned 3D shape editing task. Given an input shape and an edit description, the objective is to edit the shape to match the edit description. When language representations are entangled, this can often produce edits not localized to the relevant parts. In contrast, by regularizing the disentanglement of language representation \emph{alone}, our system achieves higher edit locality while making more pronounced edits in the requested areas.} 
  \label{fig:teaser}
\end{figure}

%% file: figures_latex/listening_overview.tex
\begin{figure*}[!ht]
  \centering
  \includegraphics[width=0.9\textwidth]{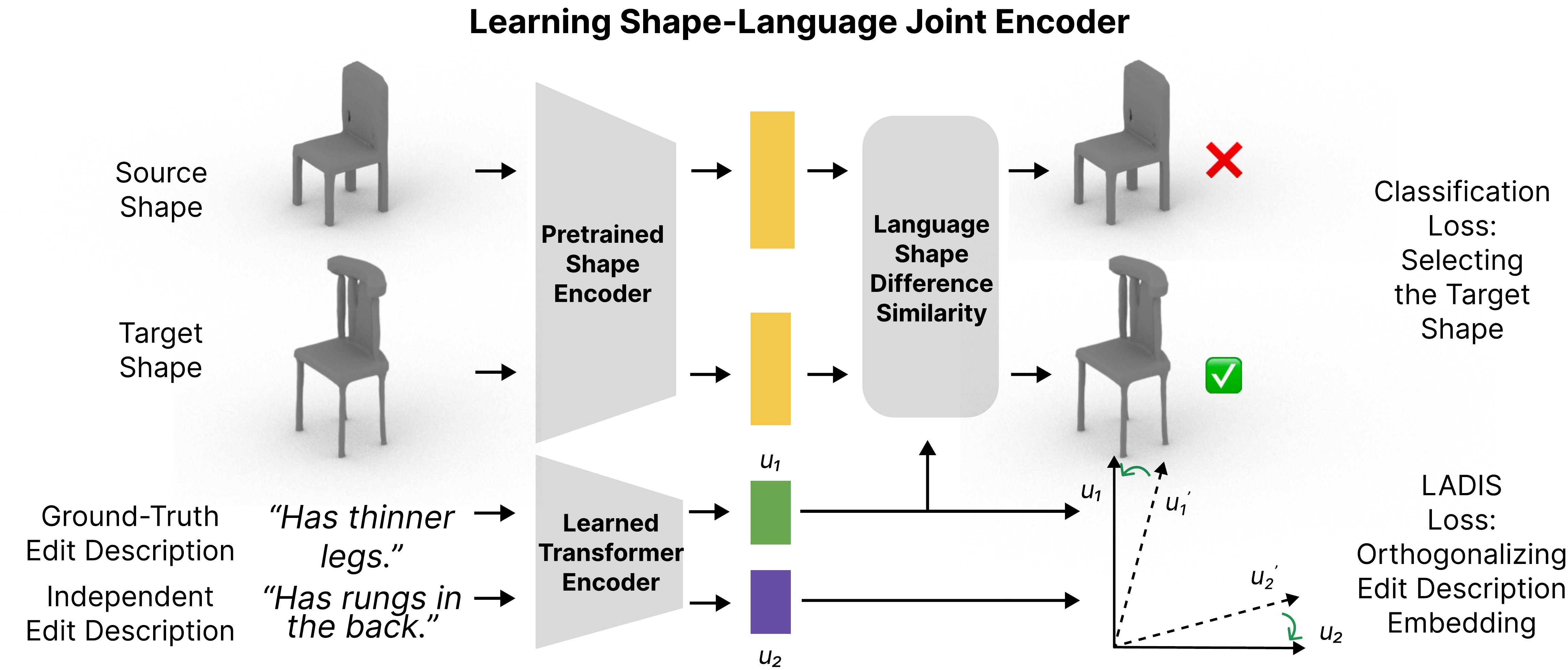}
  \caption{Overview of the training for a disentangled language-shape difference joint space. Our network solves a binary classification task of determining which input shape is the target, and in this process must jointly reason about both the shape differences and the edit description. To disentangle this space, we regularize the process with the LADIS loss, which encourages independent instructions to be orthogonalized within the joint space.}
  \label{fig:listeneroverview}
\end{figure*}

%% file: figures_latex/editing_overview.tex
\begin{figure*}[!ht]
  \centering
  \includegraphics[width=\textwidth]{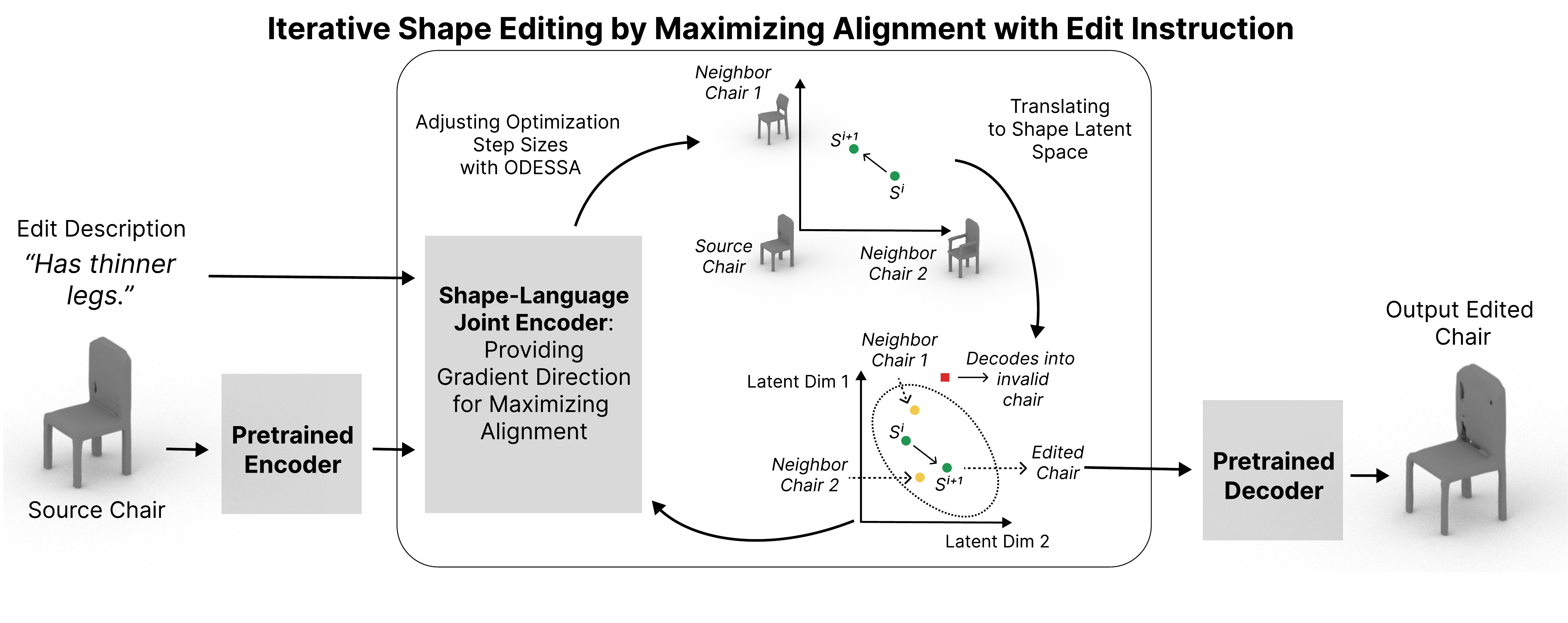}
  \caption{Overview of the editing procedure. The system first retrieves a set of nearest neighbors to create an coordinate space in which our edited shape will be expressed. Using the joint space $\mathcal J$, gradient steps iteratively increase the semantic similarity of the edited shape  difference and the edit description, modulated by ODESSA. The pretrained decoder can be used to decode the final edited latent code to an output.}
  \label{fig:editing_overview}
\end{figure*}

%% file: figures_latex/listening_architecture.tex
\begin{figure*}[!ht]
  \centering
  \includegraphics[width=0.9\textwidth]{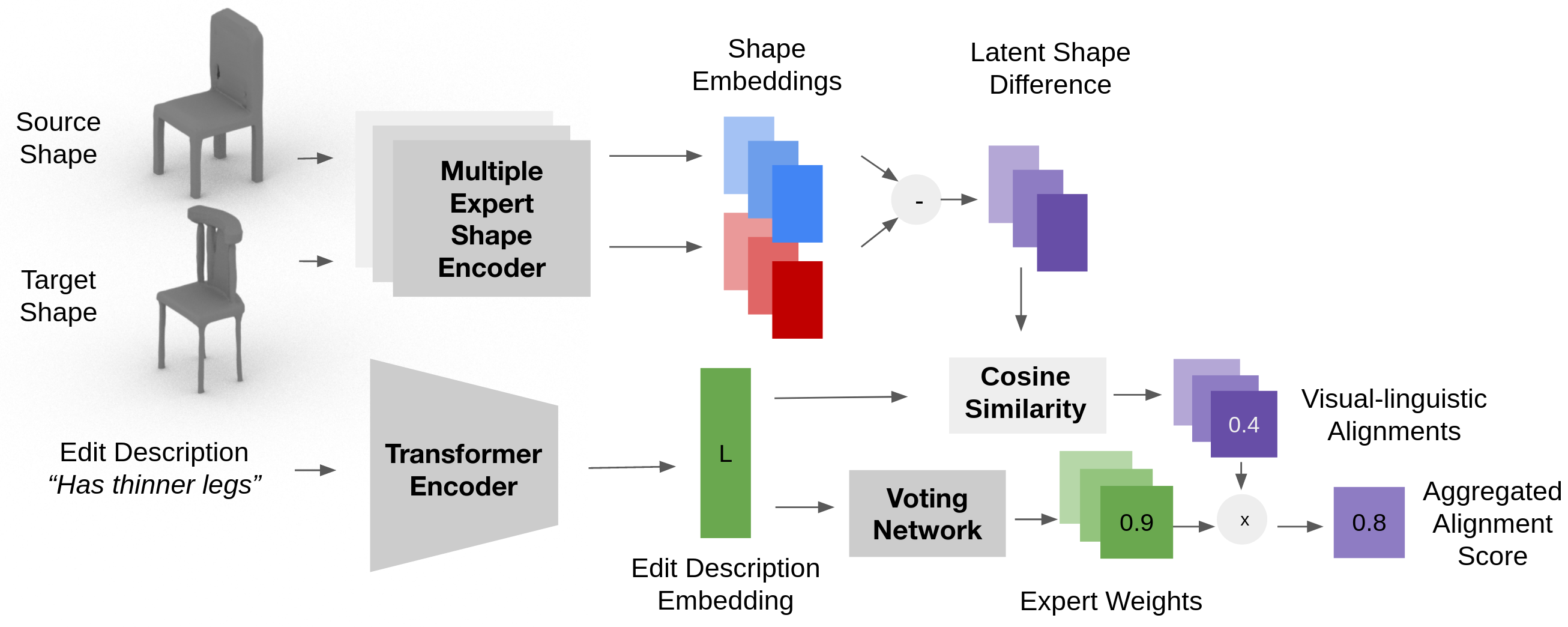}
  \caption{The architecture used to learn a disentangled joint space for language and shape differences. Note that multiple expert models are used to represent shape differences in different spaces, and the corresponding similarity predictions are aggregated by a voting network that chooses among these spaces based on the sentence embedding.}
  \label{fig:listening_architecture}
\end{figure*}

%% file: sections/3_related_works.tex
\subsection{Language Grounding in 3D shapes} 
Prior works~\citep{shapeglot, changeit3d, snare} study language grounding in the \textit{geometry} of common 3D objects by first collecting natural referential language, extracted by deploying referential games~\cite{Lewis69_Convention,referit} that use 3D shapes as visual grounding. Then, tapping on such visio-linguistic data the aforementioned works and similar ones~\cite{achlioptas2020referit_3d,zhenyu2019scanrefer,3d_Scent, PartGlot}, train deep-learning-based models by learning to solve 3D-grounded referential tasks. 

While our work focuses on learning disentangled representations to edit shapes, we also show that our proposed method can lead to improvements in language reference resolution as well.
ChangeIt3D~\citep{changeit3d} introduced the language-driven shape editing task and contributed a large scale dataset (ShapeTalk) to enable data-driven solutions for it. Here, we improve upon baseline methods introduced in ChangeIt3D.
Methods like DreamFusion~\cite{poole2022dreamfusion}, DreamFields~\cite{jain2022zero}, CLIPNerf~\cite{wang2022clip}, Text2Mesh~\cite{text2mesh} and AvatarCLIP~\cite{hong2022avatarclip} differentiably render 3D objects into images and use pretrained vision-language models such as CLIP~\citep{clip} or pretrained text-to-image models like Imagen~\cite{saharia2022photorealistic} to both linguistically ground and optimize the geometry to semantically align with the language input.

While these are promising approaches, their value is mainly in generating new shapes from scratch, where the idea of ``edit locality'' no longer applies. Nonetheless, the question of how to produce minimal, localized edits to a shape remains an important one, if text-guided shape editing is to be widely adopted in the future for 3D content creation pipelines.
As such, our work studies grounding language in pure 3D representations, and specifically aims to produce \emph{disentangled} shape edits.

\subsection{Disentangled Representations}
In the deep learning literature, disentangled representations have been studied mostly for learning generative models. InfoGAN~\cite{infogan} pioneered this line of research while modifying a GAN objective function to maximize the mutual information between latent codes and the generated data. $\beta$-VAE~\cite{betavae} extended this idea with a regularization loss enforcing statistical independence of the latent factors, which enabled more stable training and the use of fewer assumptions on the data distribution. Follow-up work such as CorEx~\cite{corex}, $\beta$-TCVAE~\cite{betatcvae}, and FactorVAE~\cite{factorvae} proposed information-theoretic approaches to resolve this issue with a total correlation penalty, and a generalization of the mutual information metric for multivariate cases. HFVAE~\cite{hfvae} also modified the objective function to achieve hierarchical factorization, and DIP-VAE~\cite{dipvae} introduced a different regularization based on covariance penalization.

For 3D shape editing, there is little work on learning disentangled representations. Aumentado Armstrong et al. ~\cite{Aumentado:2019} were the first to leverage the ideas of HFVAE~\cite{hfvae} and DIP-VAE~\cite{dipvae} on 3D shape data. The approach was extended in DeepMetaHandles~\cite{deepmetahandles} to learn keypoint-based intuitive meta-handles. DeformSyncNet~\cite{deformsyncnet} also proposed learning a dictionary for editing, whose disentanglement is enforced with a sparsity regularization over the per-point offsets of the deformation. Compared to these works, we introduce a novel method for learning disentangled \emph{language} representation to edit 3D shapes given a language prompt.

%% file: sections/4_method.tex
\subsection{Problem Statement}
We are interested in the language-based shape editing task: 
given a source shape $S$ and edit description $u$, we want to modify $S$ to fulfill the edits mentioned in $u$.
Our model learns to accomplish this task by learning on a dataset of edit descriptions describing the \emph{difference} between some source shape $S$ and target shape $T$.
Importantly, as shown in \Cref{fig:teaser}, $u$ is not expected to describe \textit{all} the characteristics of either $S$ or $T$, just some ways in which $T$ is \textit{different} from $S$.
$S$ and $T$ can be given to us in various 3D representations (meshes, implicits, etc.).
We assume access to pretrained shape autoencoders that can both encode to and decode from latent vector representations of shapes. 
In other words, $s=\textit{enc}(S)\in \mathbb R^{d}$ is a vector representing the 3D shapes $S$ and $S=\textit{dec}(s)$ represents the inverse operation.

The key to tackle the language-based shape editing task is linking an edit instruction $u$ with the shape difference between the target $T$ and source $S$, which we denote as $\textit{diff}(S, T)$. 
We achieve this by embedding $\textit{diff}(S,T)$ and $u$ into the same vector space $\mathcal{J}$, as other works like Text2Shape~\citep{chen2018text2shape} and CLIP~\citep{clip} have done. 

Inspired by prior work studying shape differences~\citep{mo2020structedit}, we represent $\textit{diff}(S,T)$ as first-class citizens. Specifically, we model $\textit{diff}(S,T) \in \mathcal J$ as $f(t) - f(s)$, where $f(x) \in \mathcal J$ is a shape encoder on top of the encoded latent representation $x$ of the input shape. To project the edit instruction to $\mathcal J$, we use encoder $g(u) \in \mathcal J$. We learn $f$ and $g$ so that the cosine similarity $\text{cossim}(f(t)-f(s), g(u))$ is high.
Equipped with $\mathcal{J}$, we then can perform editing using optimization. We initialize $s'$ as a perturbed version of $s$ and iteratively change $s'$ based on $u$ to maximize $\text{sim}(f(s')-f(s), g(u))$. 
During editing, we keep the parameters of $f, g$ unchanged and only keep updating $s'$.
We refer these two stages as \textbf{Grounded Representation Learning} and \textbf{Optimization-based Editing}.
We now discuss these two stages respectively.

\subsection{Grounded Representation Learning}
In order to learn a good grounded representation space $\mathcal J$, we propose to learn a classification task.
Given the triplet $(s, t, u)$, we train the encoders $f$ and $g$ by learning a binary classification task of identifying whether $s$ or $t$ is the correct edit target, 

\begin{align*}
\mathcal L_{\text{binary}}&(s, t, u) = \\
& -\log \biggr ( \frac{\exp h(s, t, u)} {\exp{h(s, t, u)} + \exp{h(t, s, u)}} \biggr )  
\end{align*}

\noindent where $h(s, t, u) =  \text{sim}(f(t) - f(s), g(u))$.

While $L_{\text{binary}}$ is useful for learning grounded representations, it does not encourage disentangling the representations in $\mathcal J$.
In order to obtain a disentangled joint space $\mathcal J$, we need to disentangle the representations of both the shape and the edit instruction.
As shown in \Cref{fig:listeneroverview}, we first design a multi-expert network architecture that separates each shape representation into multiple specialized vectors and then propose a language disentanglement loss that orthogonalizes representations of independent edit instructions.
In the next two sections, we discuss these two techniques respectively.

\subsubsection{Multi-Expert Shape Difference Encoder}
There are many different perspectives on describing a 3D shape, including dimension, texture, structure, etc.
Ideally, we can represent each aspect of a shape with a different representation.
We instantiate this intuition by designing a multi-expert architecture (\Cref{fig:listening_architecture}), where a collection of experts $f_1, f_2 ... f_k$ project the input shapes into separate spaces.
For an expert $f_i$, we define the shape difference as $f_i(t) - f_i(s)$. We aggregate each of the projected shape differences via a voting network $w(u) = \text{Softmax}(MLP(g(u)))$, by:

\begin{equation}
f(t) = \sum_{i=1}^k w_i(u)f_i(t).
\end{equation}

With $f$ as a mapping from each shape into $\mathcal J$, we measure the alignment $h(s, t, u)$ between the shape difference and edit instruction by cosine similarity with the encoding of the edit instruction representation $g(u)$ ,
\begin{equation}
h(s, t, u) = \text{cossim}(g(u), f(t)-f(s)).
\end{equation}


In practice, we implement $g$ with a transformer encoder~\cite{attentionisallyouneed} and take the representation of the first token. We implement each shape encoder expert $f_i$ as an MLP ontop of the latent shape representation from a pretrained shape autoencoder.
We additionally make the temperature parameter $\tau$ for $w(u)$ a learnable parameter, which empirically lowers throughout training automatically as the experts specialize.



    
\subsubsection{Language Disentanglement Loss} \label{sec:ladisloss}



Recall that our end goal of learning a disentangled joint space $\mathcal J$ is to ensure localized edits: when the user asked for making the legs thinner, the system should not add holes to the back of the source chair (~\Cref{fig:teaser}).
In our framework, this intuition translates to ensuring that independent edit instructions $u$ and $u^{-}$ have zero similarity, i.e. $\text{cossim}(g(u), g(u^{-})) \approx 0$.

With this goal in mind, we design a regularization loss by mining independent edit instructions.
We use $ M(u) = \{u^{-}_1, u^{-}_2, ...u^{-}_k\} $ to represent the action of retrieving independent instructions for $u$.
With these independent edit instructions, we directly optimize for our previous intuition by proposing a language disentanglement (LADIS) loss,
\begin{equation}
\mathcal L_{\text{LADIS}} (u) = \sum_{u^{-} \in M(u)} \lvert g(u) \cdot g(u^{-}) \rvert.
\end{equation}


In practice, we design different sampling strategies for mining independent edit instructions based on the dataset structure.
We optimize the shape difference encoder by adding $\mathcal L_{LADIS}$ and $\mathcal L_{\text{binary}}$ and tune the ratio between them as a hyperparameter.

\subsection{Shape Editing with Iterative Optimization}\label{sec:editing_procedure}
While a disentangled shape-language joint space $\mathcal J$ provides strong signals, we now discuss the algorithm to convert these signals to actual shape edits.
Recall that we encode the input shape $S$ using a pretrained auto-encoder into $s$.
As shown in \Cref{fig:editing_overview}, we follow the standard approach to iteratively traverse the latent encoder space to find a new shape $s'$ that maximizes $h(s, s', u)$ and then decoded $s'$ with the pretrained decoder (\Cref{alg:editing}). 

However, the encoder latent space is nonlinear and a naive optimization often leads to degenerate edits.
To better guide the latent space traversal, we propose two modifications: we adjust the direction of traversal by leverage a valid object neighborhood and adjust the traversal step size by controlling for the total change incurred by the edit in the output.

\begin{algorithm}[t] 
\caption{Calculate edited latent $s'$}
\begin{algorithmic}[1]
\REQUIRE input shape $s$, edit utterance $u$,
\REQUIRE shape decoder $\textit{dec}$, Number of nearest neighbors $P>0$, Variance $\gamma>0$, Expected output change $\delta > 0$, $B$ optimization steps.
\STATE $ Q \leftarrow$ GetNearest$(P, s) $
\STATE $ \epsilon \sim N(0, \gamma) $
\STATE $ s' \leftarrow s + \epsilon^T Q $
\FOR{$B$ steps} 
\STATE $ \Delta \epsilon \leftarrow \nabla_{\epsilon} h(s, s', u) $
\STATE $ \eta \leftarrow \textit{ODESSA}(s', \Delta \epsilon, \delta)$ 
\label{line:nearest}
\STATE $ \epsilon \leftarrow \epsilon + \eta \Delta \epsilon$
\STATE $ s' \leftarrow s + \epsilon^T Q$
\ENDFOR
\RETURN $s'$
\end{algorithmic}
\label{alg:editing}
\end{algorithm}


\subsubsection{Neighborhood Simplex Editing (NSE)} \label{sec:NSE} 

To prevent the latent space traversal from deviating off the valid shape manifold, we retrieve $P$ nearest neighbors from the training set using $s$, forming a $P$-simplex $Q \in \mathbb{R}^{P\times d}$ within a higher-dimensional latent space (\Cref{alg:editing}, Line 6).
We constrain the traversal by treating $Q$ as base coordinates, and optimizing $s'$ with these new coordinates by setting $s' = s + \epsilon^{T}Q$, as shown in \Cref{fig:editing_overview}.
Our intuition is that $P$ can be chosen well enough to cover most modes of variations from the source shape $s$, while being low-dimensional enough (in comparison to the latent space dimension $d$) to constrain the editing to happen on the shape manifold. 
Note that these modes of variation do not necessarily have to be captured in a ``positive'' directional sense. For example, since the elements of $\epsilon$ are allowed to be negative, for the utterance ``shorter legs'', a neighboring chair within the chair space with \textit{longer} legs than the input shape can be equally as informative as one with \textit{shorter} legs.
Ablations involving NSE is included in the Appendix.

\subsubsection{Output-Driven Edit Step-Size Adjustment (ODESSA)} \label{sec:ODESSA}

The high-dimensional nonlinear nature of the encoder latent space makes editing challenging because a small step in the latent space can potentially cause both catastrophically large changes as well as unnoticeable changes in the decoded shape.
To combat this, we rescale the size of each edit step so that the total change in the decoded output is similar.
We denote this operation as $\textit{ODESSA}(s', \Delta \epsilon, \delta)$, which outputs a scalar as the step size in the direction $\Delta \epsilon$ based on an expected total change $\delta$ in the output space.
We implement \textit{ODESSA} by calculating the changes on the decoded 3D representation and present the implementation details and relevant ablations in the Appendix.

%% file: sections/5_experiments.tex
In this section, we evaluate our system's ability to achieve localized language-guided shape edits.
We evaluate our disentangled joint space $\mathcal J$ by shape classification accuracy, and by the locality of our system's edits, evaluated both by our novel part-based metric and human evaluators.
In addition, we analyze expert specialization in our multi-expert shape difference encoder and visualize the disentangled language representations.
Experimental results show that our proposed method is better at shape classification, produces more disentangled representations, and lead to more local shape edits.

\subsection{Experiment Setup}

\subsubsection{Datasets}

There are two datasets that provide textual information for differences between shapes: ShapeTalk~\cite{changeit3d} and ShapeGlot~\cite{shapeglot}. In both, text describes a distinguishing feature that a target shape has which the source shape(s) do not. Within ShapeTalk, each sample has only a single source shape, but multiple separate lines of text are collected from human labelers that enumerate many differences between the source and the target. For ShapeGlot, the text is more pragmatic, and used to differentiate a target from \textit{two} source shapes. This also means that a single textual description can very sparsely describe the difference between the source and the target. In our experiments, we use these textual descriptions as edit descriptions.


\subsubsection{Evaluating Metrics} \label{sec:pep}
\input{sections/5a_evalmetric}

\subsection{Experimental Results}

\input{figures_latex/qualitative_samples}

\subsubsection{Shape Editing}

We investigate whether the disentangled language representations allow for more localized edits. We compare against the SOTA method~\citep{changeit3d} on the ShapeTalk dataset using mPEP.

\paragraph{Latent Geometric Representations}
In our editing experiments, we use the IM-NET~\cite{chen2019learning} as a shape autoencoder, as it enables easy conversion to meshes, and therefore enables better comparisons of edit locality.

\paragraph{Mining for independent utterances}
For ShapeTalk, labelers are prompted to enumerate different shape differences for every source-target pair. As such, the utterances provided for a single pair by a single labeler are very likely independent. We can thus instantiate $M(u)$ as the  multiple utterances gathered from a single labeler for the same source-target pair (\textbf{multiutterance LADIS}). As a compromise in settings where labeler information is not available,  we instantiate $M(u)$ as the other edit instructions within the dataset used to describe the \textit{same source-target pair} (\textbf{shared-context LADIS}). 

\paragraph{Experimental Results}
Table \ref{tab:compare2changit3d} compares our method with ChangeIt3D. We find that LADIS not only  produce edits that are 20\% more localized (measured by mPEP), but also volumetrically larger changes ($\geq 3 \times$ that of ChangeIt3D edits), allowing our system to achieve more visually pronounced and localized shape edits than ChangeIt3D edits, as shown in the sample in \Cref{fig:teaser}. 
We perform a dependent sample T-test between the ChangeIt3D baseline and our method (specifically the multi-utterance variant) and confirm that the PEP improvement was statistically significant (p=0.0287).
This verifies our hypothesis that learning disentangled representations can better ensure the locality of the geometric edits, even though no explicit part information was ever given to the model.

\input{figures_latex/shapetalk_editing_qualitative}

\input{tables/shapetalk_editing_compareChangeIt3D}

\subsubsection{Language Reference Resolution}
We next measure the quality of the shape-language joint space by evaluating the source-target shape binary classification accuracy, which is the training objective of shape difference encoder.

\paragraph{Latent Shape Representations}
In addition to IM-NET, we experiment with the ShapeGF~\cite{cai2020learning} autoencoder, which learns to reconstruct shapes by deforming an initial prior distribution of points according to a gradient field.

\input{tables/shapetalk_accuracy}

\paragraph{Mining for independent utterances}
As an additional compromise where the set of utterances per source-target is prohibitvely sparse, we explore using utterances from different source-target pairs within the batch to be $M(u)$ (\textbf{random LADIS}). We do this with ShapeGlot, and break every triplet of shapes into two pairs $(\text{source1},\text{target})$ and $(\text{source2},\text{target})$ for training and testing.

\input{tables/shapeglot_accuracy}

\paragraph{Experimental Results}
 \Cref{tab:shapeglot_listening_acc} and \Cref{tab:shapetalk_listening_acc} show  that models trained with the LADIS loss has a notable increase in classification accuracy across two datasets and two backbone networks. 
 \Cref{tab:shapeglot_listening_acc} also reveals that even when loosely independent edit instructions are used for LADIS loss, we still observe a performance gain on the ShapeGlot dataset. 
 Our findings show that LADIS produces higher quality language-shape joint spaces $J$, which may explain its effectiveness for downstream shape editing.

\subsubsection{Human Evaluation}

\Cref{fig:qualitative_samples} displays some examples of our method's shape edits on the chair category. Qualitative results show that on average, our method produces more semantically accurate and localized edits.

To evaluate this more systematically, we conduct a human evaluation comparing our method and ChangeIt3D.
Our evaluation employs 10 volunteer annotators from our institute and in total, has 500 annotations.  
Our annotators are given the input, a language description of the desired output, and a pair of edited outputs (from our method and ChangeIt3D), and are tasked with deciding which of the edits: (1) is most obviously aligned with the language instruction and (2) is the most localized (i.e. preserves the identity of the unmentioned parts of the object).
More details about the annotation process is detailed in the Appendix.

In 72\% of the cases, our edits were deemed more semantically accurate (more semantically aligned and visually obvious) compared to ChangeIt3D. In 61\% of the cases, our results were deemed more localized.  Looking across the evaluators, all 10 evaluators unanimously judge that our method produces more accurate and obvious edits than ChangeIt3D. 8 of the 10 evaluators judge that our method produces strictly more localized edits than ChangeIt3D, with the other 2 evaluators indicating a tie.

In 65\% of the cases, a higher PEP score directly corresponds to a subjective perception of higher locality of edits reported by human evaluators. This also motivates the use of PEP as an automatic evaluation metric.

\subsubsection{Visualizing Disentanglement}

\input{figures_latex/language_embedding}

What is the effect of the LADIS loss on the language representations? Figure \ref{fig:language_embedding} shows a T-SNE embedding of a subset of edit descriptions within the joint space $\mathcal J$.
We color-code edit description based on the various chair parts that are mentioned. 
Comparing representations trained with LADIS to a baseline without LADIS, we see that representation clusters are more distinct and better correspond to certain part identities. 

To analyze the effect of the LADIS loss on the shape representations, we visualize expert specialization.
We group edit descriptions by the adjective that appear in them and aggregate expert weights $w(u)$.
Recall that expert weights $w(u)$ decides the importance of each expert representation for different edit instruction.
\Cref{fig:expertactivation} shows experts trained with LADIS loss specialized to different adjectives.
For example, whenever an edit description contains the word ``longer'', the seventh expert is weighted most heavily.
In contrast, without the LADIS loss, expert specialization occurs to a much lesser extent.
In conclusion, our representation analysis suggests that the LADIS loss helps disentangle both the language and the geometry representations.

\input{figures_latex/shapetalk_expert_distribution}


%% file: sections/5a_evalmetric.tex
In order to quantify  the degree to which our edits are local to certain parts, we introduce \textbf{Part-based Edit Precision} (PEP), which is defined as the log of the ratio between the percentage volume change found in the ``relevant" regions of the input 3D shape, versus the entire shape. For a subset of chairs, we manually associate certain keywords within edit descriptions to Partnet~\citep{mo2019partnet} part labels, and this enables us to measure the percentage volume change observed in the regions associated with mentioned parts. The PEP score is $0$ when the percentage volume change in the mentioned area is the same as the percentage volume change across the whole entire shape. Thus, higher PEP indicates that an edit has high locality. We additionally measure $\Delta v$, the total change in volume. Further information about PEP and $\Delta v$ is provided in the Appendix.

In what follows, we evaluate the mean PEP (mPEP) score and mean $\Delta v$ (m$\Delta v$) for 2000 chairs within the ShapeTalk test set with existing PartNet~\citep{mo2019partnet} part segmentations.



%% file: figures_latex/qualitative_samples.tex
\begin{figure*}[!ht]
  \centering
  \includegraphics[width=\textwidth]{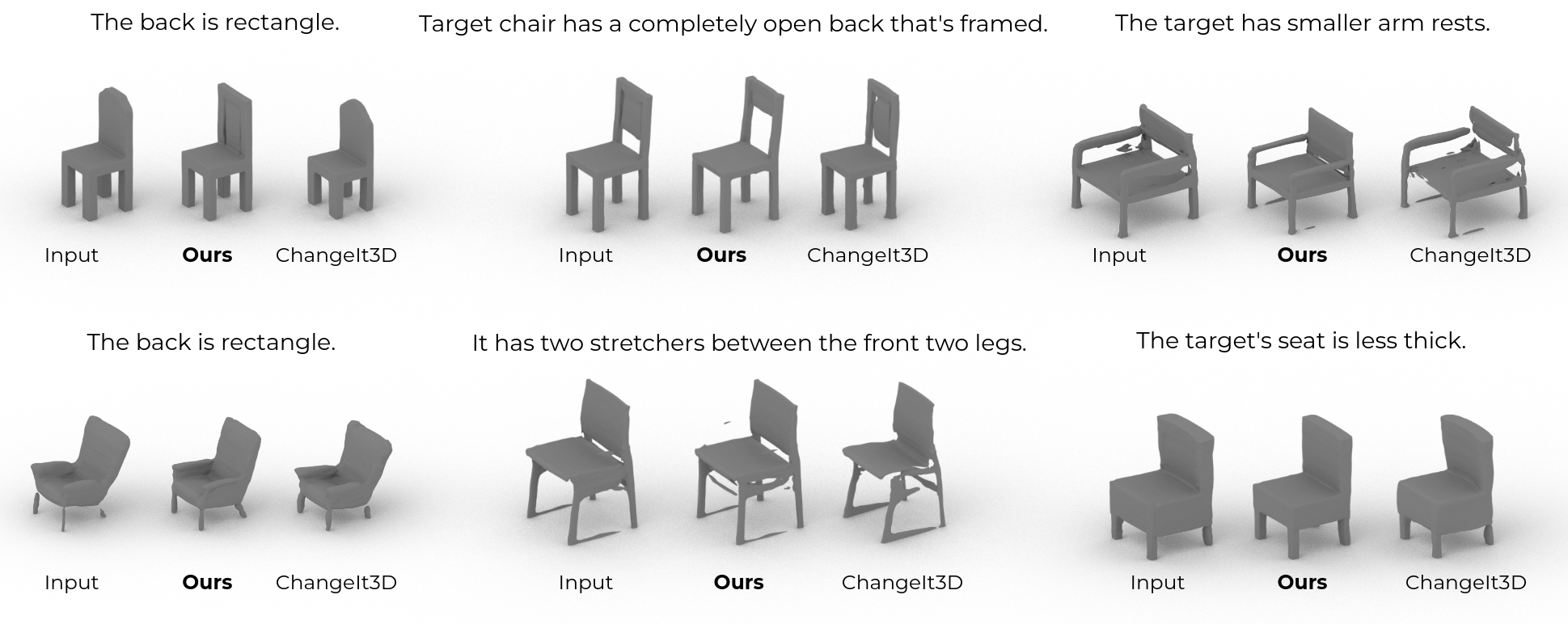}
  \caption{Qualitative examples of our method editing different kinds of chairs, compared to ChangeIt3D. The text \emph{describing the desired output} is shown above the samples. Compared to ChangeIt3D, our method allows for more semantically correct and obvious edits, while improving the locality of the shape edit.}
  \label{fig:qualitative_samples}
\end{figure*}

%% file: figures_latex/shapetalk_editing_qualitative.tex

%% file: tables/shapetalk_editing_compareChangeIt3D.tex
\begin{table}
\centering
\begin{tabular}{cccc}
\hline
\ & Multiutt & ShareCon &  ChangeIt3D \\
\hline
mPEP $\uparrow$ &  \textbf{0.378} & 0.338 &  0.315 \\
m$\Delta$V $\uparrow$  & 0.0334 & \textbf{0.0416} & 0.0105 \\
\hline
\end{tabular}
\caption{Comparison of edit locality (mPEP) and mean edit volume (m$\delta$V) with the ChangeIt3D baseline on ShapeTalk. For multiutterance LADIS (multiutt) and shared-context LADIS (ShareCon), our models achieve higher locality and more volumetrically obvious edits.}
\label{tab:compare2changit3d}
\end{table}

%% file: tables/shapetalk_accuracy.tex
\begin{table}
\centering
\begin{tabular}{cccc}
\hline
\ & Multiutt & ShareCon &  ChangeIt3D \\
\hline
IM-NET & \textbf{69.52\%}  & 69.03\% & 62.9 \% \\
ShapeGF & 70.62\%  & \textbf{70.94\%} & 65.64 \% \\ 
\hline
\end{tabular}
\caption{Source-target classification accuracies of multiutterance LADIS (Multiutt) and shared-context LADIS (SharCon) on ShapeTalk, compared to ChangeIt3D \cite{changeit3d}. Both versions of our models outperform ChangeIt3D, on both types of latent representations.}
\label{tab:shapetalk_listening_acc}
\end{table}

%% file: tables/shapeglot_accuracy.tex
\begin{table}
\centering
\begin{tabular}{cccc}
\hline
\ & Random LADIS & w/o LADIS \\
\hline
IM-NET & \textbf{78.43\%}  & 76.15\% \\
ShapeGF & \textbf{76.58\%}  & 75.59\% \\ 
\hline
\end{tabular}
\caption{Source-target binary classification accuracies on ShapeGlot. LADIS loss applied on even random utterances yield an notable improvement in accuracy for both IM-NET and ShapeGF latent representations.}
\label{tab:shapeglot_listening_acc}
\end{table}

%% file: figures_latex/language_embedding.tex

\begin{figure}[!ht]
  \centering
  \includegraphics[width=\columnwidth]{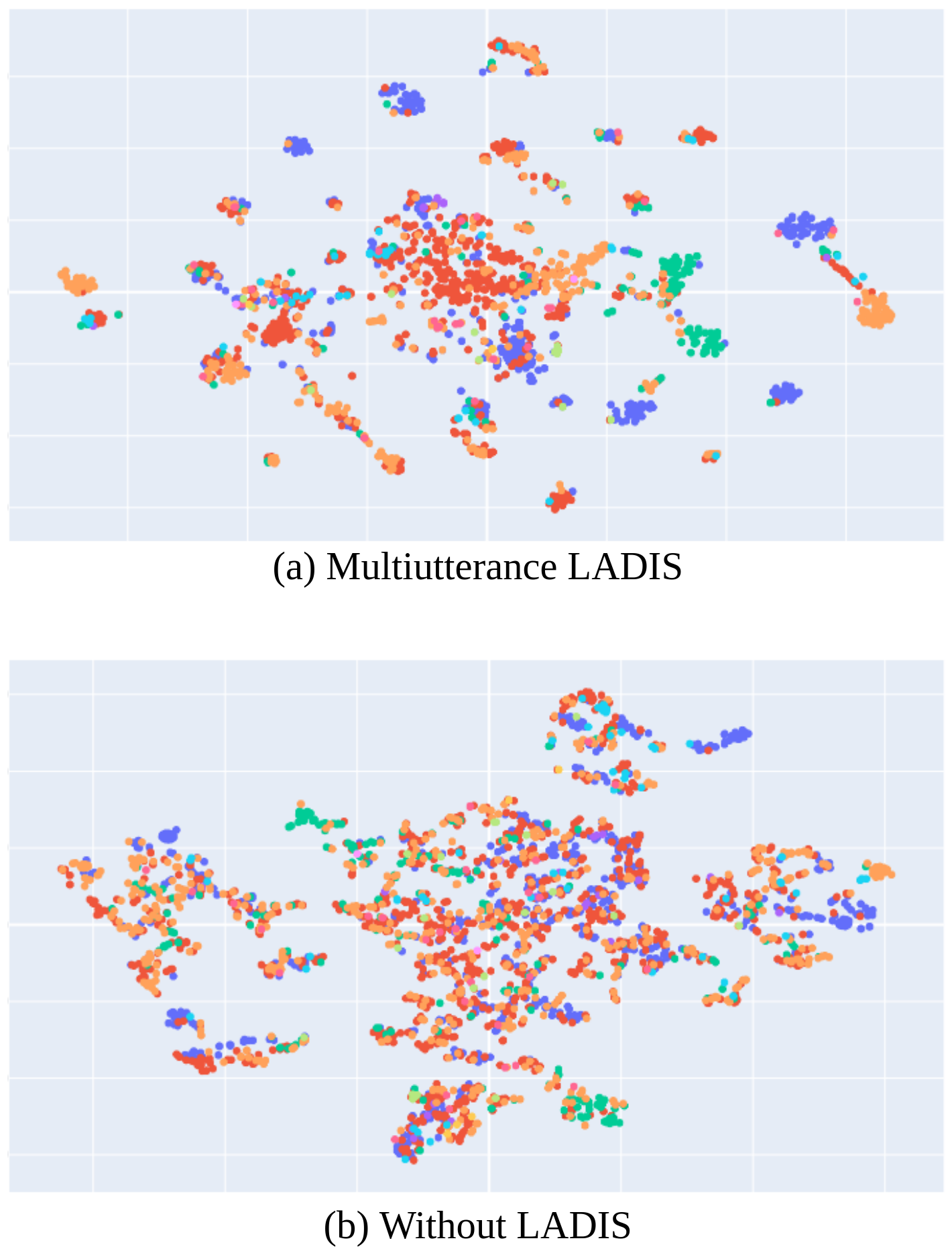}
  \caption{TSNE visualization of the learned language embeddings for \textbf{(a)} multi-utterance LADIS and \textbf{(b)} our model without LADIS loss, with the colors indicating the sets of parts mentioned within each utterance.}
  \label{fig:language_embedding}
\end{figure}

%% file: figures_latex/shapetalk_expert_distribution.tex
\begin{figure}[!t]
  \centering
  \includegraphics[width=0.70\columnwidth]{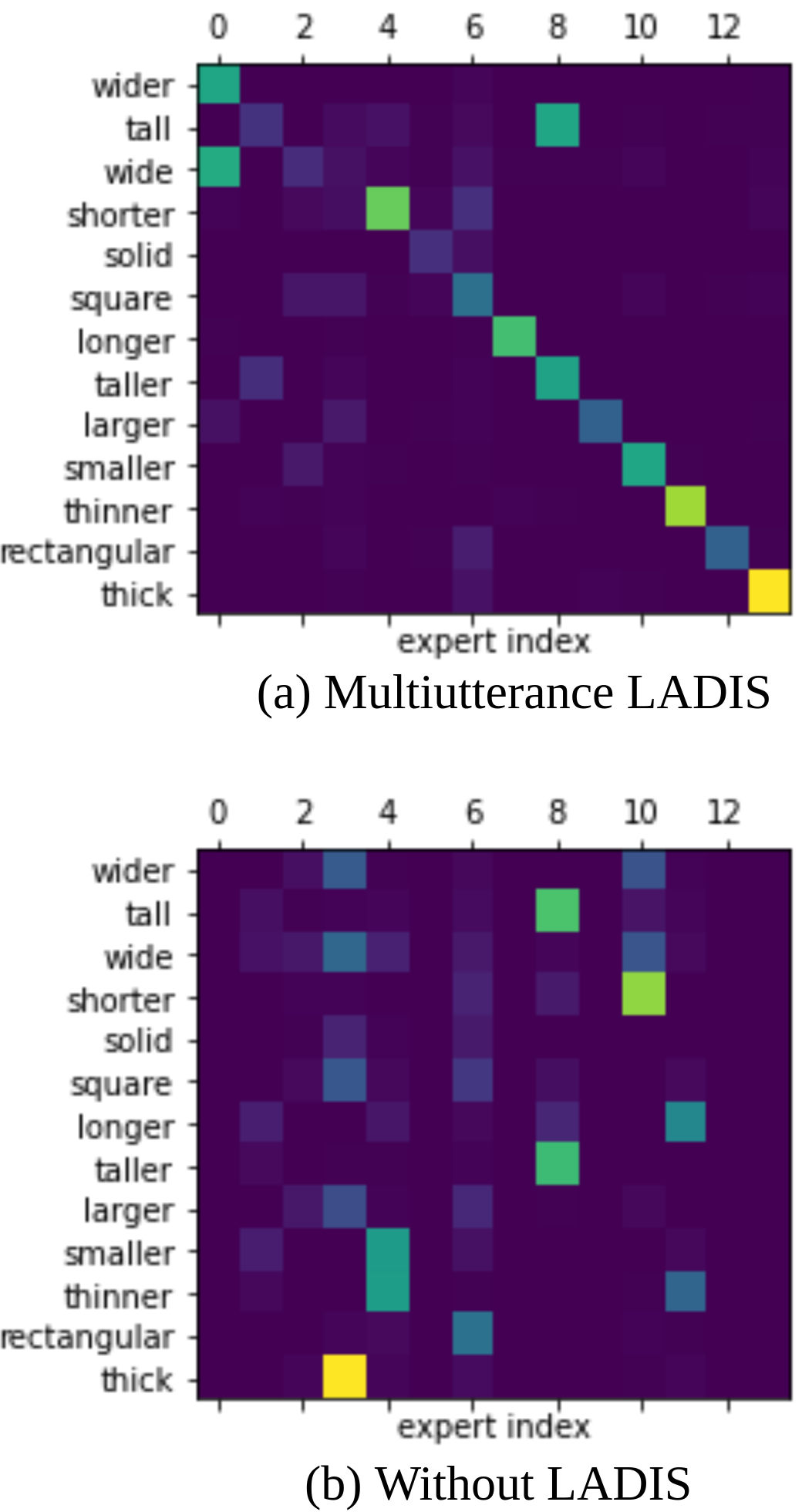}
  \caption{Visualizing the activation of different experts for different utterances reveals that the LADIS loss allows for different experts activate for different sets of adjectives. Notice that synonymous words like ``wider''/``wide" and ``tall''/``taller'' are allocated to the same experts. Compare to the baseline without LADIS loss, where the same level of specialization is not as clearly observed.} 
  \label{fig:expertactivation}
\end{figure}

%% file: sections/6_conclusion.tex
In this work, we introduce a set of complimentary techniques to improve language-based shape edits.
Our core idea is to learn a more disentangled shape-language joint space, which is achieved by a novel multi-expert architecture and a language disentanglement loss. Importantly, we show that we can accomplish more obvious edits while improving their locality by learning to disentangle the language \emph{alone}, without ever equipping our model with part supervision.
Experimental results show the effectiveness of our methods in producing localized shape edits.
Our techniques are agnostic to the underlying 3D representations, and we look forward to extending our method to other 3D representations and modalities. 

%% file: sections/8_appendix.tex
In this appendix we will provide further detail on the implementation of our models, and the parameters during training and editing. We will also provide the mathematical definition of Part-wise Edit Precision (PEP), and the specific implementation of Output-Driven Edit Step-Size Adjustment (ODESSA). Through ablations, we will demonstrate the importance of language disentanglement for the task of localized shape edits, and also the importance of our edit-time design decisions (namely, ODESSA and Neighborhood Simplex Editing). Finally, we'll provide further details on the human evaluation conducted to evaluate our method.

\subsection{Implementation details}

When learning the joint space, we train using batchsizes of 64, with 256-dimensional encodings for every shape (for both ShapeGF and IM-NET autoencoders). We pretrain the shape autoencoders and the joint spaces on the official ShapeTalk training set. For both IM-NET and ShapeGF, we use the default training hyperparameters, matching the pretraining process of autoencoders in ChangeIt3D. For both autoencoders, the latent space is 256 dimensional.

The dimensionality of our model's joint space is 128 dimensional, and use 14 expert networks, each with 3 feedforward layers with skip connections. For the implementation  of the transformer, we use a slightly modified version of the text encoder used for CLIP with 4 layers and 4 multi-attention heads, which produces a sentence embedding that is 1024 dimensional at the first token.

The voting network is implemented as a single fully connected layer, with the associated softmax temperature parameter (for producing weights of experts) initialized at 1.0 during training. We set the relative weight between the binary classification loss and LADIS loss to be 1 -- that is, our training loss is $\mathcal L = \mathcal L_{binary} + \mathcal L_{LADIS}$. Training is done for 20 epochs and the checkpoint with the highest validation accuracy on the official ShapeTalk validation set is selected.

When editing using the pretrained joint space, the number of neighbors we choose for NSE is 64. We also use gradient descent for 50 steps, with each step rescaled according to ODESSA.

\input{tables/shapetalk_editing_languageAblations}

\subsection{Definition of Part-Wise Edit Precision}

Part-Wise Edit Precision is a novel metric we introduce to measure the locality of an edit to the parts that are explicitly mentioned by language instructions. The metric measures the log of the ratio between percentage  volume change found in the correct regions of the input 3D shape versus the entire shape. 

Let sequence of edits to a shape be $s(t)$, where $s(0)$ is the original input shape. We assume that for input shape $s_0 = s(0)$, we have a part decomposition $R_{s_0}: \mathbb R^3 \rightarrow P$ that maps a given point in $\mathbb R^3$ to a label within $P$, the full set of part labels. We can construct $R_{s_0}$ by using part-segmentation masks from PartNet.

We also construct a mapping $Q: L \rightarrow \mathcal P(P)$, which is a manually designed classifier that, based on the words within the utterance $u$ from space $L$, selects a subset of $P$ as ``relevant'' parts, where changes should be allowed. We define $Q$ manually by mapping sets of words found in utterances of ShapeTalk to different PartNet part classes.


Given an implicit occupancy decoder $\textit{dec}$, we can find the volume difference $\Delta v$ defined for some edit $s' = s(t)$, and some region $X$:

\begin{equation}
\Delta v(X, s') = \biggr |\  \int_{x \in X} \textit{dec}(s', x)  - \textit{dec}(s_0, x) dx \ \biggr |
\end{equation}

We can then find the percentage volume change in region $X$ as:
\begin{equation}
w(X, s') = \frac{\Delta v(X,s')}{\int_{x \in X} \textit{dec}(s_0, x) dx} \,.
\end{equation}

PEP is defined as the log of the ratio of percentage volume change found in the parts of the input object mentioned by language to the percentage volume change found in the whole object:

\begin{equation}
PEP(s(t), u) = \log \frac{w(\mathcal X_{s(0), u}, s(t) )}{w(\mathcal X, s(t))}\,,
\end{equation}
    
\noindent where $\mathcal X$ is the ambient space and $\mathcal X_{s(0), u} = \mathcal X \cap R^{-1}_{s(0)}(Q(u))$ is the set of points that belong in the set of parts mentioned by utterance $u$. When an edit is more "local" in a part-wise sense, we expect PEP to be high. Since  the ratio is established between two percentage changes, this effectively normalizes against the size of the part mentioned, which rewards nominally smaller volume changes to skinny/small parts mentioned by language if it amounts to a significant percentage of the original part volume. We find empirically that this is important, as many utterances refer to volumetrically less-significant parts, like legs and armrests.

We've also found empirically that the $\log$ of the ratios has a much more Gaussian-like distribution, which matches the human prior we assume over visual obviousness of a change in the correct parts. 

Due to the use of part segmentation masks (used to define $R_{s_0}$), PEP is susceptible to errors in the masks, as well as translations and rotations within edits that may shift parts far outside the groundtruth masks. The latter may be mitigated by measuring and comparing edits of smaller magnitude, or “swelling” the groundtruth masks, as we have done in our implementation. Additionally, the PEP metric depends on a predefined set of part semantic groups (e.g. for chairs, we’ve used “back”, “seat”, “legs” and “armrests”), and is therefore specific to a fixed part semantic group for each object category.

However, since PEP does not have to rely on a pre-trained part classifier (as done in the metric proposed by ChangeIt3D) and accounts for the differences in size of parts by assigning volumetrically smaller parts higher weight, we believe that it is still a stronger metric than what currently exists. PEP therefore should be used to do system-level comparisons of edit locality.


To represent the size of the edit, we measure the mean volume change m$\Delta$v at edit timestep $t$ as the mean of $v(\mathcal X, s(t))$ over different (source shape $s_i(0)$, language $u$) pairs. Over the same set, we measure the mean PEP at a certain edit timestep $t$ as the mean over all edits:
\begin{equation}
mPEP = \frac{1}{N} \sum_{i=1}^N PEP(s_i(t), u_i)
\end{equation}

Note that while the above is formulated for implicit occupancy autoencoders like IM-NET, it can be extended to other types of autoencoders by other notions of edit distance, such as chamfer distance between surfaces.

\input{tables/shapetalk_editing_editablations}

\subsection{Implementing ODESSA}

ODESSA rescales the size of each step to combat the problem that latent space traversal often produces unpredictable differences in the  space, which can be especially detrimental when we optimize within the joint space over multiple optimization steps.

Thus, ODESSA rescales the size of each step according to some desired degree of total change in the decoded output per optimization step, denoted by the parameter $\delta$:

\begin{align*}
\textit{ODESSA} (s', \Delta \epsilon, \delta) = \\
\delta \times \biggr | \biggr ( \nabla_s \int_{x \in \mathbb R^3 } \text{dec}(s, x) dx \biggr ) ^T ( \Delta \epsilon ^T Q ) \biggr |^{-1}
\end{align*}

Note that in the above formulation, we assume that the geometry is outputted by the decoder as an implicit representation, which means allows the integral over the 3D space to capture total variation within the output occupancy values. But as is the case with PEP, the following formulation can also be extended to other autoencoder architectures by using other notions of edit distance like chamfer distance between surfaces.

\subsection{Ablations}

In Table \ref{tab:edit_language_ablations}, we compare the effect of gradually relaxing the assumption of language independence on the improvement of the edit locality across multiple edits. We do these experiments on the ShapeTalk dataset, which supports the strongest assumptions of edit description independence (i.e. multiutterance LADIS). We find that the better the negative set is (i.e. stronger independence), the higher the mPEP performance. This suggests that the degree to which we can mine for independent utterances heavily influences the degree of edit locality, and furthermore that this trend holds throughout iterative edits.

\input{figures_latex/nse_ablation}
In Table \ref{tab:editprocedureablations}, we compare the effects of removing NSE and ODESSA. Specifically, since these two techniques are meant to preserve the shape class and also produce more obvious edits to shapes, we compare the Frechet Inception Distance (FID) of the renderings of decoded edits to that of the distribution of the decoded input latent codes. We find that without using NSE, our model's FID largely increases, while we notice a slight decrease in the mean amount of occupancy difference in the output shape. This suggests that without using NSE, edits easily shift the edited shape off of the valid shape manifold (See \Cref{fig:nse_ablate}). On the other hand, without using ODESSA to recalibrate the optimization step sizes, the mean amount of occupancy difference (m$\Delta$V) becomes comparable to the ChangeIt3D baseline (see results in \Cref{tab:compare2changit3d}) -- that is, the edits become measurably less obvious.


These ablations therefore suggest that our two modifications to the canonical optimization-based editing procedure both allow for more obvious edits while improving the ability of our edits to stay on the valid shape manifold.

\input{figures_latex/shapetalk_editing_ablations_qualitative}

\subsection{Human Evaluation}
\input{figures_latex/human_evaluator_interface}
\Cref{fig:human_interface} shows the interface we used to collect human evaluation data to evaluate our qualitative results. We asked 10 human evaluators to each annotate 50 sample edits (500 annotations total), comparing our method and ChangeIt3D. 8 of the 10 annotators have no experience with NLP, graphics or machine learning. No time limit was enforced on the annotation process. The average time spent on annotations was 35 minutes.

Each evaluator is given a 2 minute introduction to the task, primarily explaining how to use the interface, and defining and providing examples for ``edit locality'' and ``edit correctness''. Importantly, evaluators are asked to comparing edits in \emph{both} criteria.

Because some of the 3D edits are fairly subtle, the "toggling functionality" enables the user to quickly flip back and forth between the source and target shapes. We found this to be very useful in deciding edit locality and correctness of relatively smaller or thinner parts (e.g. legs of chairs.).

We also found that for structural changes  (e.g. ``the chair has stretchers''), human evaluators typically have a harder time deciding which one is more obvious or accurate if both (or neither) edits being compared achieve the structural change. As such, we allow for ties in the correctness/obviousness criterion.


%% file: tables/shapetalk_editing_languageAblations.tex
\begin{table*}[!ht]
\centering
\begin{tabular}{ccccc}
\hline
Step &  Multiutt LADIS &  ShareCon LADIS &  Random LADIS &  w/o LADIS \\
\hline
1 &           \textbf{0.372} &           0.345 &      0.271 &     0.185 \\
2 &           \textbf{0.384} &           0.320 &      0.217 &     0.191 \\
3 &           \textbf{0.373} &           0.316 &      0.252 &     0.182 \\
4 &           \textbf{0.321} &           0.303 &      0.252 &     0.171 \\
5 &           \textbf{0.305} &           0.269 &      0.241 &     0.162 \\
\hline
\end{tabular}
\caption{Different variants of LADIS, from most ideal to least, from left to right. mPEP shown for 5 edit steps, where the output of the previous step is the input shape into the next step, for the same edit description. We see that mPEP generally decreases from the most to least ideal variant of LADIS.}
\label{tab:edit_language_ablations}
\end{table*}

%% file: tables/shapetalk_editing_editablations.tex
\begin{table}[!ht]
\centering
\begin{tabular}{cccc}
\hline
\ & Multiutt & w/o ODESSA &  w/o NSE \\
\hline
m$\Delta$V & 0.0334 & 0.0095 &  0.0274 \\
FID & 33.58 & 21.36 & 46.04 \\
\hline
\end{tabular}
\caption{Ablations in different ablations in the edit-time procedure. We verify through this that removing the ODESSA step results in less obvious volumetric changes, and that without NSE, our method produces less plausible edit outputs.}
\label{tab:editprocedureablations}
\end{table}

%% file: figures_latex/nse_ablation.tex
\begin{figure}
  \centering
  \includegraphics[width=0.7\columnwidth]{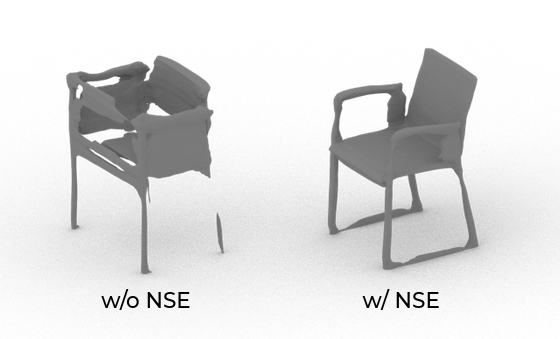}
  \caption{An example of the same optimization-based editing, run with and without NSE. Note how without NSE, the optimization produces spurious artifacts that degenerates the chair output.}
  \label{fig:nse_ablate}
\end{figure}

%% file: figures_latex/shapetalk_editing_ablations_qualitative.tex

%% file: figures_latex/human_evaluator_interface.tex
\begin{figure*}[!ht]
  \centering
  \includegraphics[width=0.9\textwidth]{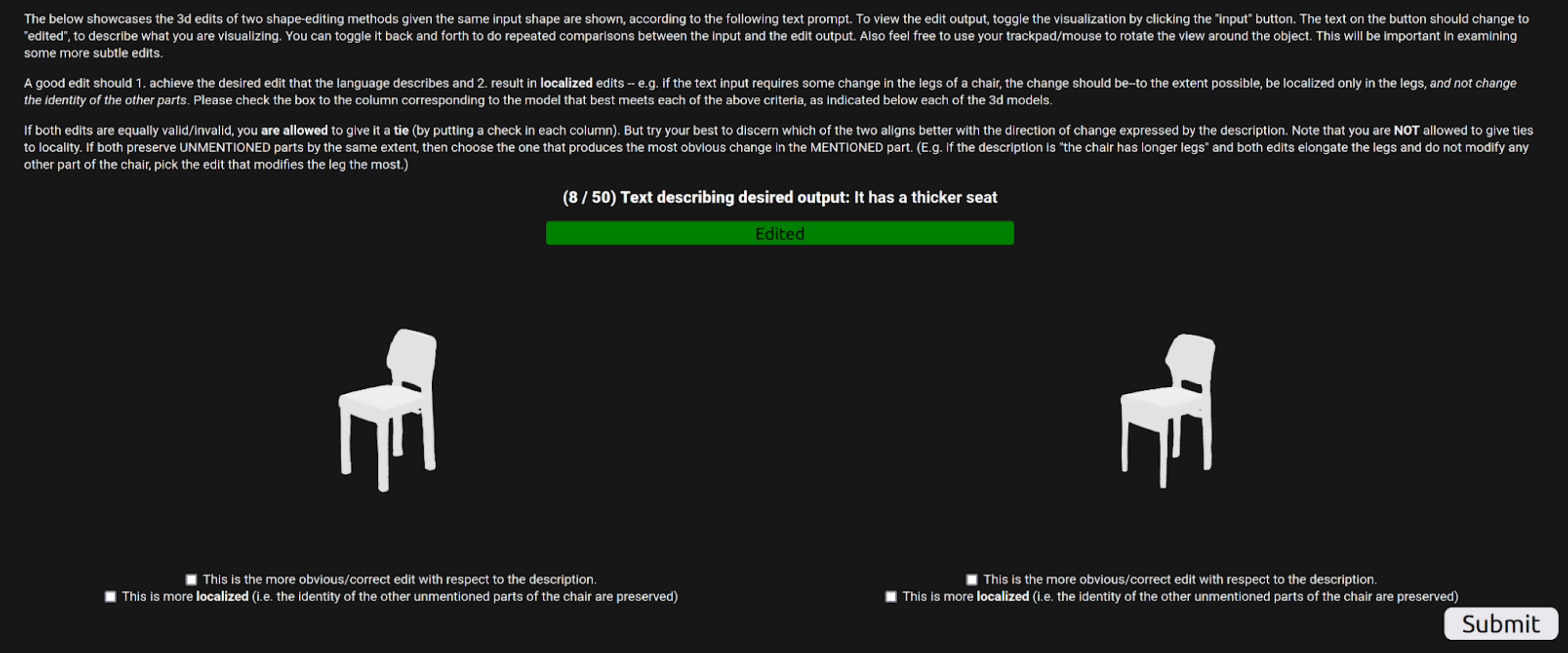}
  \caption{The web interface used by human evaluators to annotate the quality of shape edits. We display the two edits side by side, one of which is produced by ChangeIt3D, and the other from our method. Through repeatedly clicking the button in green, we allow users to toggle the view from inspecting the \emph{input} shape to the edited \emph{output}, so that minute differences in edits (e.g., slight scaling of certain parts) of both methods can be more easily detected than side-by-side comparisons. Users can also rotate the 3D models around, as well as zoom in/out, to inspect the influence of the edit on different parts of the input. Before moving onto the next sample, the evaluator must select which edit is more semantically obvious/correct, and which is more localized. We clearly define what each of these criteria mean in the instructions.}
  \label{fig:human_interface}
\end{figure*}